\documentclass[11pt, twoside, onecolumn, letterpaper]{article}

\RequirePackage{wsc}
\usepackage{latexsym}
\usepackage{graphicx}
\usepackage{mathptmx}
\usepackage[T1]{fontenc}

\usepackage[pdftex,colorlinks=true,urlcolor=blue,citecolor=black,anchorcolor=black,linkcolor=black]{hyperref}

\usepackage{booktabs, dcolumn, makecell, mathtools, orcidlink, siunitx}
\usepackage[inline]{enumitem}
\usepackage[capitalise,noabbrev]{cleveref}
\newcolumntype{T}{>{\ttfamily}l<{}}
\newcolumntype{C}{>{$}c<{$}}

\renewcommand\*[1]{^{\textnormal{#1}}}

\usepackage{tikz}
\usetikzlibrary{positioning,arrows.meta}
\definecolor{Blue}{RGB}{0,115,189}
\definecolor{Green}{RGB}{0,134,131}
\pgfdeclarelayer{low}
\pgfdeclarelayer{middle}
\newcommand\curvebox[3][0]{%
     \filldraw[line width=.5mm, fill=white, line join=round] (-#2,-#3+#1) -- (-#2,#3+#1) .. controls (-.3*#2,#3+#1+.2) and (.3*#2,#3+#1-.2) .. 
     (#2,#3+#1) -- (#2,-#3+#1) .. controls (.3*#2,-#3+#1-.2) and (-.3*#2,-#3+#1+.2) .. cycle;
}

\tikzset{
  module/.style={
    align=center,
    font=\color{white}, 
    line width=2pt,
    draw=white, 
    fill=#1!60
  },
  module/.default=Blue,
  func/.style={
    align=left,
    font=\color{white}, 
    line width=2pt,
    draw=white,
  },
  func/.default=Blue,
  bg/.style ={
    align=center, 
    font=\color{black}, 
    line width=2pt,
    draw=#1!20, 
    fill=#1!10
  },
  bg/.default=Blue,
  lab/.style ={
    draw=none, 
    fill=none, 
    anchor=south west,
    minimum width=0, 
    minimum height=0cm
  },
  small/.style ={
    text width=3.2cm, 
    minimum height=1.1cm
  },
  tiny/.style ={
    text width=2.75cm, 
    minimum height=5mm 
  },
  head/.tip={Stealth[width=2.5mm, length=2.5mm]},
  shead/.tip={Stealth[width=1mm, length=1mm]},
  arrow/.style={
    ->,
    rounded corners,
    line width=.5mm,
    black,
    > = head
  },
  ddash/.style={
    -,
    densely dashed,
    line width=.3mm
  },
  pics/file/.style args={#1+#2}{code={
    \pgfmathsetmacro\w{1.35}
    \pgfmathsetmacro\h{1.4}
    \pgfmathsetmacro\ybox{-.6}
    \pgfmathsetmacro\wbox{\w-.2}
    \curvebox{\w}{\h}
    \curvebox[\ybox]{\wbox}{.5}
    \node [anchor=center] at (0, \h-.5) {#1};
    \foreach \dh in {1.55, 1.8}
        \pgfmathsetmacro\y{-\h+\dh}
        \draw[line width=.4mm] (-\w+.3,\y) .. controls (-.3*\w,\y+.2) and (.3*\w,\y-.2) .. (\w-.3,\y);
    \node [anchor=center, text width=1.2*\wbox cm, text centered] at (0, \ybox) {\baselineskip=0pt\small #2\par};
  }},
}

\usepackage[normalem]{ulem}
\colorlet{revise}{blue}
\newcommand\add[1]{\textcolor{revise}{#1}}

\DeclareRobustCommand\remove{\bgroup\markoverwith{\add{\rule[.5ex]{2pt}{1pt}}}\ULon}
\DeclareRobustCommand\hl{\bgroup\markoverwith{\textcolor{revise!20}{\rule[-.5ex]{.1pt}{2.5ex}}}\ULon}

\begin{document}


\pagestyle{fancyplain}

\thispagestyle{plain}
\firstPageHead{}

\chead{\fancyplain{}{\itshape Strand, Gorton, Asprusten, and Brathen}}

\rhead{}
\cfoot{}
\renewcommand{\headrulewidth}{0pt} 

\makeatletter
\let\@internalcite\cite
\def\cite{\def\@citeseppen{-1000}%
    \def\@cite##1##2{(##1\if@tempswa , ##2\fi)}%
    \def\citeauthoryear##1##2##3{##1 ##3}\@internalcite}
\def\citeNP{\def\@citeseppen{-1000}%
    \def\@cite##1##2{##1\if@tempswa , ##2\fi}%
    \def\citeauthoryear##1##2##3{##1 ##3}\@internalcite}
\def\citeN{\def\@citeseppen{-1000}%
    \def\@cite##1##2{##1\if@tempswa, ##2)\else{}\fi}%
    \def\citeauthoryear##1##2##3{##1 (##3)}\@citedata}
\def\citeA{\def\@citeseppen{-1000}%
    \def\@cite##1##2{(##1\if@tempswa , ##2\fi)}%
    \def\citeauthoryear##1##2##3{##1}\@internalcite}
\def\citeANP{\def\@citeseppen{-1000}%
    \def\@cite##1##2{##1\if@tempswa , ##2\fi}%
    \def\citeauthoryear##1##2##3{##1}\@internalcite}
\def\shortcite{\def\@citeseppen{-1000}%
    \def\@cite##1##2{(##1\if@tempswa , ##2\fi)}%
    \def\citeauthoryear##1##2##3{##2 ##3}\@internalcite}
\def\shortciteNP{\def\@citeseppen{-1000}%
    \def\@cite##1##2{##1\if@tempswa , ##2\fi}%
    \def\citeauthoryear##1##2##3{##2 ##3}\@internalcite}
\def\shortciteN{\def\@citeseppen{-1000}%
    \def\@cite##1##2{##1\if@tempswa, ##2\else{}\fi}%
    \def\citeauthoryear##1##2##3{##2 (##3)}\@citedata}
\def\shortciteA{\def\@citeseppen{-1000}%
    \def\@cite##1##2{(##1\if@tempswa , ##2\fi)}%
    \def\citeauthoryear##1##2##3{##2}\@internalcite}
\def\shortciteANP{\def\@citeseppen{-1000}%
    \def\@cite##1##2{##1\if@tempswa , ##2\fi}%
    \def\citeauthoryear##1##2##3{##2}\@internalcite}
\def\citeyear{\def\@citeseppen{-1000}%
    \def\@cite##1##2{(##1\if@tempswa , ##2\fi)}%
    \def\citeauthoryear##1##2##3{##3}\@citedata}
\def\citeyearNP{\def\@citeseppen{-1000}%
    \def\@cite##1##2{##1\if@tempswa , ##2\fi}%
    \def\citeauthoryear##1##2##3{##3}\@citedata}
%
%
%
\def\@citedata{%
    \@ifnextchar [{\@tempswatrue\@citedatax}%
                  {\@tempswafalse\@citedatax[]}%
}

\def\@citedatax[#1]#2{%
\if@filesw\immediate\write\@auxout{\string\citation{#2}}\fi%
  \def\@citea{}\@cite{\@for\@citeb:=#2\do%
    {\@citea\def\@citea{, }\@ifundefined
       {b@\@citeb}{{\bf ?}%
       \@warning{Citation `\@citeb' on page \thepage \space undefined}}%
{\csname b@\@citeb\endcsname}}}{#1}}%

%
\def\@citex[#1]#2{%
\if@filesw\immediate\write\@auxout{\string\citation{#2}}\fi%
  \def\@citea{}\@cite{\@for\@citeb:=#2\do%
    {\@citea\def\@citea{; }\@ifundefined
       {b@\@citeb}{{\bf ?}%
       \@warning{Citation `\@citeb' on page \thepage \space undefined}}%
{\csname b@\@citeb\endcsname}}}{#1}}%

%
\def\@biblabel#1{}
\makeatother



\newdimen\bibindent
\bibindent=0.0em
\def\thebibliography#1{\section*{\refname}\list
   {}{\settowidth\labelwidth{[#1]}
   \leftmargin\parindent
   \itemindent -\parindent
   \listparindent \itemindent
   \itemsep 0pt
   \parsep 0pt}
   \def\newblock{}
   \sloppy
   \sfcode`\.=1000\relax}


\setlength{\baselineskip}{12.7pt}

\title{LEARNING ENVIRONMENT FOR THE AIR DOMAIN (LEAD)}

\author{Andreas Strand\\
    Patrick Gorton\\
    Martin Asprusten\\
    Karsten Brathen \\[12pt]
    FFI\\
	$\left(\begin{tabular}{@{}c@{}}Norwegian Defence\\ Research Establishment\end{tabular}\right)$\\
	Instituttveien 20\\
	2007 Kjeller, NORWAY
}
\maketitle

\section*{ABSTRACT}
A substantial part of fighter pilot training is simulation-based and involves computer-generated forces controlled by predefined behavior models. The behavior models are typically manually created by eliciting knowledge from experienced pilots, which is a time-consuming process. Despite the work put in, the behavior models are often unsatisfactory due to their predictable nature and lack of adaptivity, forcing instructors to spend time manually monitoring and controlling them. Reinforcement and imitation learning pose as alternatives to handcrafted models. This paper presents the Learning Environment for the Air Domain (LEAD), a system for creating and integrating intelligent air combat behavior in military simulations. By incorporating the popular programming library and interface Gymnasium, LEAD allows users to apply readily available machine learning algorithms. Additionally, LEAD can communicate with third-party simulation software through distributed simulation protocols, which allows behavior models to be learned and employed using simulation systems of different fidelities.

\section{INTRODUCTION}\label{sec:intro}
A large part of the training fighter pilots undergo occurs in simulators under instructor supervision. In these simulators, the pilots practice tactics and operations by engaging in scenarios including friendly and hostile forces, often represented by computer-generated forces (CGFs), which are autonomous or semi-autonomous actors used in military simulation \shortcite{lovlid17}. These CGFs must behave in a way that accelerates training and builds the necessary competence of the pilots. Still, a current limitation to using CGFs for training is that their behaviors often come across as predictable, inviting pilots to exploit their vulnerabilities rather than focus on achieving the training objectives \shortcite[ch.~1]{toubman20}. Such constraints in the behavior
models force instructors to micromanage the CGFs, restricting the complexity of scenarios that can be managed and trained \shortcite{kallstrom22}. Besides, qualified instructors are both in short supply and on tight schedules, meaning they should devote their full attention to giving instructions and feedback to pilots. Modeling adaptive and intelligent air combat behavior for CGFs is thus a key challenge.

Recent research on behavior modeling revolves around the use of modern machine learning (ML) techniques, such as deep reinforcement learning, to train agents to perform various tasks in the air combat domain \shortcite{pope21,hu21,zhang20}. Freely available software libraries like Stable Baselines \shortcite{raffin21} and RLlib \shortcite{liang18} have allowed state-of-the-art ML methods to be used in a plug and play manner using the Gymnasium interface (previously Gym, \shortciteNP{brockman16}). While there are some promising results concerning learning air combat behavior, most of the agents created in these studies were tailored to work within their specific learning environments and not intended for use with simulation systems used in pilot training. As a result, their performance may suffer from a \emph{reality gap} \shortcite{souza20}, meaning the agents will fail to operate in environments that are slightly different from their learning environment. A possible way of combating this is through \textit{transfer learning}, a technique for utilizing expertise from one task or domain to benefit the learning process of another \shortcite{zhu20}. Consequently, there is a need for a flexible system where both the learning environment and assigned tasks are modular and replaceable.

This paper describes the Learning Environment for the Air Domain (LEAD), a system developed at the Norwegian Defence Research Establishment (FFI) for applying ML to air combat tasks and concepts. A major component of LEAD is the Simple Air Combat Simulation (SACS) developed at FFI and tailored for use with ML. The simulation system is designed to be used in a distributed simulation environment and to be interchangeable with other simulation systems. This makes it possible to expose agents to multiple simulations with different characteristics, hopefully making them more robust and keeping the effects of the reality gap to a minimum. The learning environment uses the Gymnasium interface, which defines a standard format for actions, states, and rewards. This should make it possible to apply any ML implementation that supports the Gymnasium interface to LEAD.

\Cref{sec:lead} presents LEAD in terms of the components and processes that make it a learning environment for agents. In \cref{sec:experiment}, the results of a reinforcement learning (RL) experiment are presented to demonstrate the potential of LEAD. \Cref{sec:discussion} include further discussion of the system and experiment results, and \cref{sec:conclusion} presents conclusions and thoughts on future improvements to LEAD. 

\section{LEARNING ENVIRONMENT FOR THE AIR DOMAIN} \label{sec:lead}
The purpose of LEAD is to use ML to create artificially intelligent pilot behavior for combat aircraft CGFs. The system is implemented as a Gymnasium environment with which an agent may interact through the Gymnasium  interface. Gymnasium is a Python programming library developed by OpenAI serving as an application programming interface (API) between ML algorithms and learning environments. The library contains a set of learning environments often used to benchmark RL and imitation learning (IL) algorithms. Gymnasium is renowned within the RL community, which has led other RL libraries such as Stable Baselines \shortcite{raffin21} and RLlib \shortcite{liang18} to integrate Gymnasium in their frameworks, making it easy for users to experiment with different ML methods and environments seamlessly.

Machine learning methods generally require large amounts of data, sometimes millions of samples, in order to learn to solve computational tasks \shortcite{dutta18}. We designed LEAD with ML in mind. Thus, rapid simulation is essential to generate enough data to solve tasks within acceptable amounts of time. Fast air combat simulation is realized by SACS, which is a simulation system developed for LEAD. Furthermore, since LEAD follows the high level architecture (HLA) standard for distributed simulation, it is possible to replace SACS with other third-party simulation applications \shortcite{ieee10}. While SACS is simple and computationally lightweight, other simulation software may offer extended functionality and fidelity at the cost of computational resources and time. Distributed simulation may also facilitate transfer learning, where agent policies are learned partly using SACS to rapidly acquire pilot-specific behavior before transferring to higher-fidelity environments.

\begin{figure}[bt]
	\centering
    \scalebox{.93}{\input{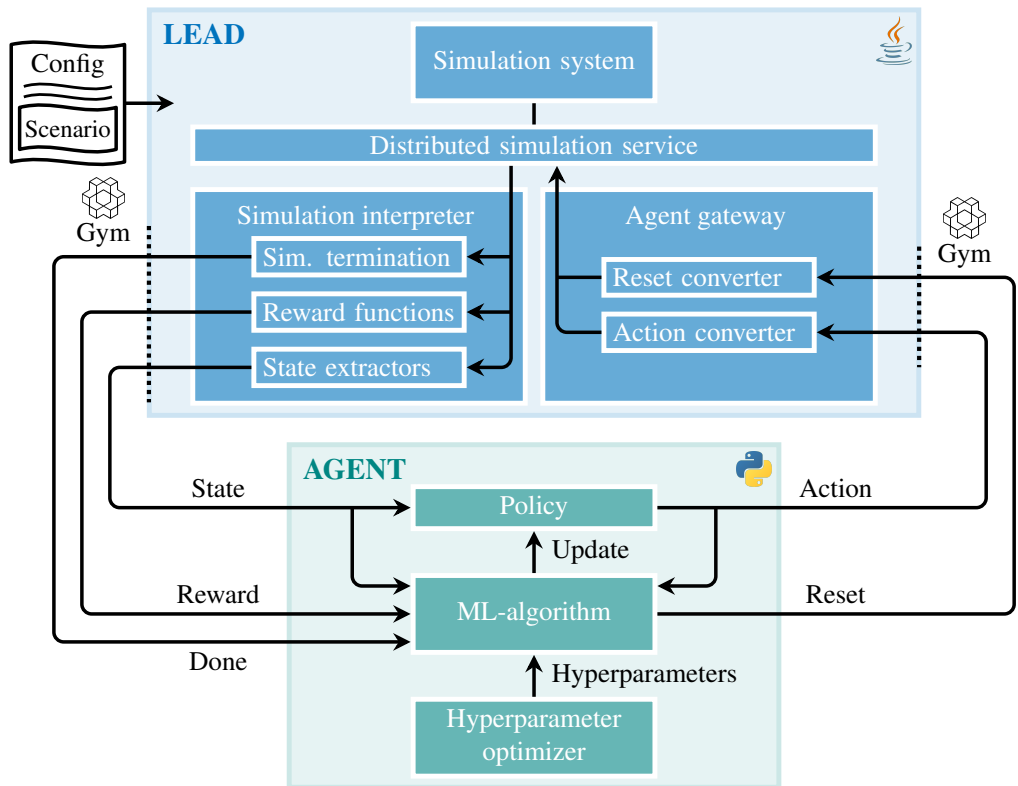}}
	\caption{An agent interacts with LEAD and learns by reinforcement or imitation. The configuration file provided to LEAD (blue section) defines the simulation, agent gateway and interpreter settings. The agent (green section) includes a selected ML algorithm and agent policy type. Learning may occur once the ML algorithm's hyperparameters are set manually or through an optimizer.}
	\label{fig:lead}
\end{figure}

The Learning Environment for the Air Domain includes
\begin{enumerate*}[label=(\arabic*), itemjoin={,\ }, itemjoin*={, and\ }]
    \item a simulation system
    \item a gateway allowing agents to control entities in the simulation system
    \item an interpreter providing agents with states and rewards
    \item a distributed simulation service that ties components \numrange{1}{3} together.
\end{enumerate*}
\Cref{fig:lead} illustrates how LEAD serves as a learning environment for agents, with particular focus on the RL process. The blue section depicts LEAD, consisting of components \numrange{1}{4}, which are implemented as Java applications and wrapped as a Gymnasium environment in Python. The green section depicts the agent, comprising a behavior policy, an ML algorithm for optimizing the policy, and the means for selecting algorithm-specific hyperparameters. These are all Python programs. The four components of LEAD are described in \cref{sec:simulation,sec:gateway,sec:interpreter,sec:connection}, while the agent and learning process are described in \cref{sec:agent} with reference to \cref{fig:lead}.

\subsection{Simulation System} \label{sec:simulation}
The Learning Environment for the Air Domain requires a simulation system to simulate entities in a virtual environment. We developed SACS to perform simple simulations in the air domain, executing faster than real time, which is an essential feature for ML. It is possible to use other simulation software in place of SACS, which may be useful for testing purposes or when simulations of higher fidelity are required. 

\begin{figure}
    \centering
    \includegraphics[width=0.99\textwidth]{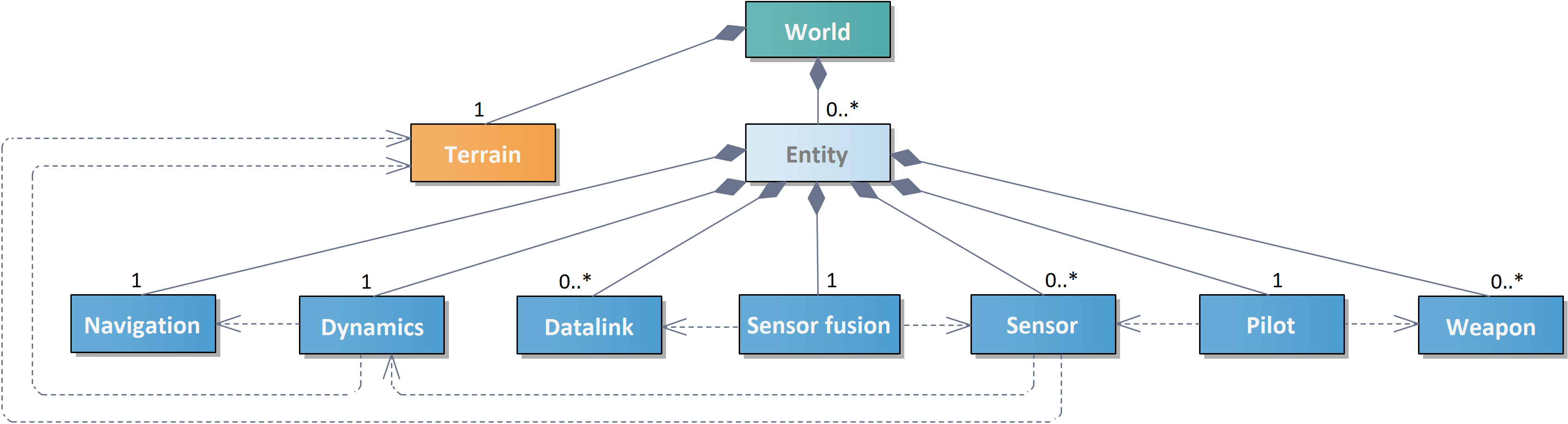}
    \caption{The architecture of SACS. The simulation system consists of a world with a terrain and entities. Each entity has seven subsystems, some of which have multiple instances or implementations. Some of these subsystems depend on other subsystems, depicted by dashed arrows.}
    \label{fig:sim_architecture}
\end{figure}

\Cref{fig:sim_architecture} presents the architecture of SACS consisting of a world with a terrain module and a collection of simulated entities. In turn, an entity has a collection of subsystems that determine how it perceives, behaves and interacts with the world. The \emph{pilot} subsystem is a core simulation component concerning ML. Simply put, the pilot subsystem can either specify behavior rules or act as a gateway between an external agent and SACS, allowing the external agent to control a simulated entity remotely. In the latter case, the pilot subsystem receives actions from the external agent via the agent gateway and applies these actions to the entity's dynamics, sensors, and weapons.

The simulation distinguishes between the ground truth state of entities and how each entity perceives the world. The navigation system generates an entity's perception of its own state, which estimates a perceived state based on its actual state in the simulation. Further, the entity may perceive the state of other entities through sensors and by receiving datalink messages from friendly entities containing state perceptions of themselves and others. The entity's sensory information and received messages are fused to avoid duplicate information and to create an understanding of the world state. In the current version of SACS, the only entity types are combat aircraft and missiles, but the architecture makes it easy to add air defense systems and more.

Each aircraft, missile, or flying object in SACS has a dynamics subsystem. It takes a set of maneuver control variables generated by the pilot subsystem as input and uses it to calculate the entity’s dynamic state. 
The maneuver control variables consist of a desired height above sea level, speed, and course that the aircraft should try to achieve. This set of variables was chosen in order to provide a suitable level of abstraction over aircraft controls, and is also similar to the type of commands accepted by other simulation systems such as VR-Forces \cite{mak21}.

The flight dynamics model (FDM) can be thought of as consisting of two main parts: an autopilot and a kinematics model. The autopilot takes the set of maneuver control variables and the aircraft's current dynamic state as inputs and uses them to calculate an acceleration and angular velocity for the aircraft. These are then passed to the kinematics model, which calculates the aircraft's next dynamic state. The aircraft's dynamic state, acceleration, and angular velocity are returned as outputs from the dynamics model.

Since ML algorithms require large amounts of data in order to learn useful behaviors, a simulation system that is capable of running much faster than real time is necessary. The kinematics model was therefore designed to run fast while maintaining an acceptable level of fidelity. To this end, many physical effects that would affect real aircraft, such as gravity and air pressure, are not considered. The physical limitations of an aircraft are approximated by placing limits on the accelerations and angular velocities the autopilot is allowed to return.

\subsection{Agent Gateway} \label{sec:gateway}
The agent takes a state as input and returns an action to the simulation. The agent and simulation do not communicate directly, so the agent gateway is needed to translate the actions of the agent into a format that is usable by the simulation system. In addition to translating actions, the agent gateway instructs the simulation to reset to its initial configuration whenever prompted by the ML algorithm. The reset functionality is essential when training ML agents because repeated runs of similar events are necessary. The agent gateway sends the controls and reset prompts to the simulation system. Details related to the distributed simulation service component are given in \cref{sec:connection}.

\subsection{Simulation Interpreter} \label{sec:interpreter}
The simulation interpreter allows agents to observe and receive feedback from the learning environment through state extractors and reward functions. These functions must be carefully designed as they greatly influence the learning process. State extractors define which aspects of the simulation the agent observes, such as the state of the controlled entity or other entities. Both perceived and ground truth states may be extracted. Reward functions measure to what extent the agent attains its goals. Each reward function returns a scalar, and these are combined to create a single reward value.

In addition to state extractors and reward functions, the simulation interpreter has a mechanism for detecting when a simulation run should terminate. If a simulation run, also called \emph{episode}, meets one or more termination criteria, the simulation interpreter announces that the simulation should reset. Through the configuration of one or multiple such state extractors, reward functions and episode termination criteria, the user defines the agent's perceptual abilities, what behavioral patterns to reinforce, and when to, if necessary, terminate an episode.

\subsection{Distributed Simulation Service} \label{sec:connection}
The distributed simulation service, or \emph{connection} for short, provides the means of communication between the simulation system, agent gateway and simulation interpreter. The flow of information in LEAD is as follows. First, the agent gateway sends its most recent controls to the respective entities in the simulation via the connection. Next, the simulation updates all simulated entities and sends these to the interpreter for state extraction and reward calculation. All data packages are stamped with a simulation time to avoid discrepancies between simulation systems, according to the principles of distributed simulation \shortcite{fujimoto99}. Currently, LEAD offers the choice between an HLA connection and an ML connection optimized for simulation speed.

\subsubsection{High Level Architecture Connection} 
The Learning Environment for the Air Domain employs a distributed simulation connection enabling interoperability between simulation systems across multiple host computers. Specifically, this connection employs HLA, which specifies the requirements of a run-time infrastructure (RTI), to which simulation systems may connect using a standard API \shortcite{ieee10,kuhl99}. The RTI is responsible for distributing simulation data and synchronizing connected systems.
\begin{figure}
    \centering
    \includegraphics[width=\linewidth]{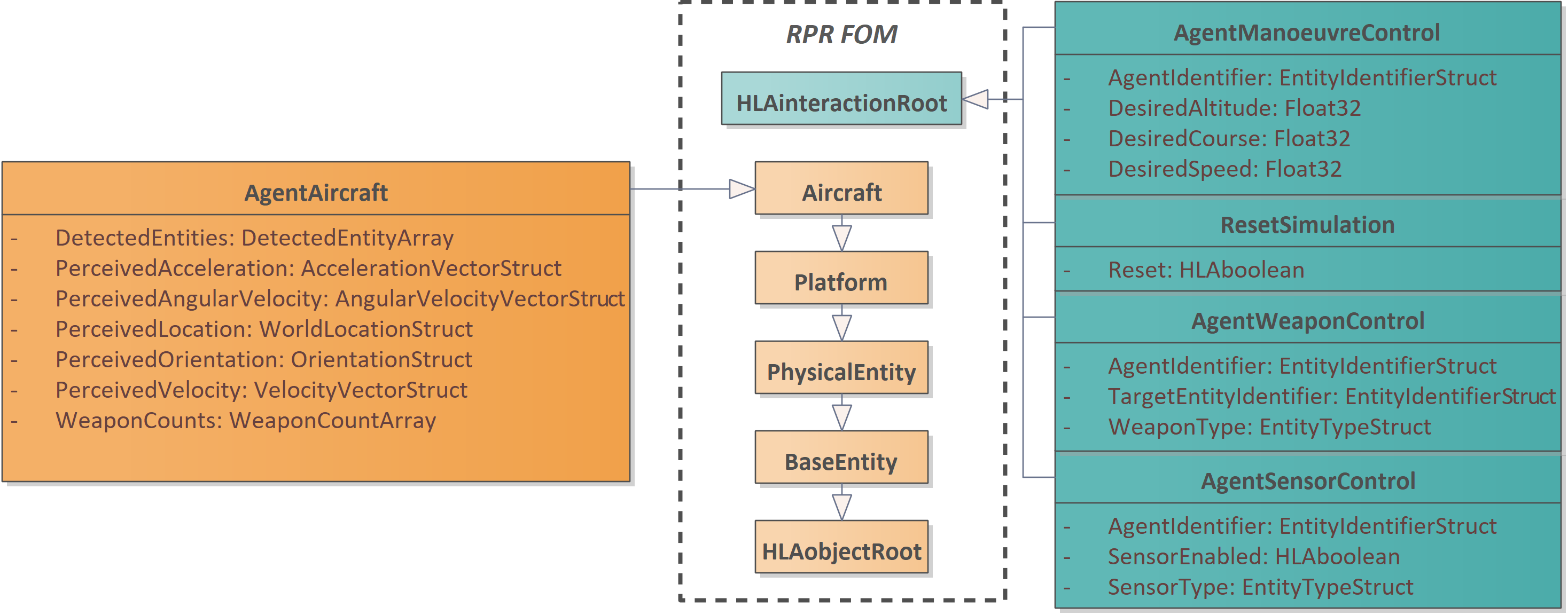}
    \caption{The HLA objects and interactions that make up the FOM used in the LEAD federation when using the HLA connection.}
    \label{fig:lead_som}
\end{figure}

When using the HLA connection, the simulation system, agent gateway, and simulation interpreter connect as \emph{federates} to a single HLA \emph{federation}. \Cref{fig:lead_som} shows the federation object model (FOM) that specifies the data model for sharing simulation data within a federation. The HLA object and interactions developed for LEAD inherit objects and interactions defined in the Real-time Platform Reference (RPR) FOM \shortcite{siso15}, shown within the dashed frame. The \emph{AgentAircraft} object, in orange, contains ground truth state attributes of an entity defined by its parent classes but is extended to contain perceived states, detected entities and weapon count. The interactions, in turquoise, contain controls that affect the simulation system.
The HLA connection also allows third-party HLA federates to connect to the federation to interact with LEAD or visualize entities simulated by SACS. Thus, it is possible to simulate complex scenarios by connecting multiple federates. This is useful for dividing the computational effort among multiple computers and allowing remote operators to join the scenario.

\subsubsection{Machine Learning Connection}
The ML connection is a simple transmission control protocol (TCP) connection intended to allow LEAD to run as fast as possible to save time during the training of agents. It is only used to connect SACS, the simulation interpreter and the agent gateway. It uses a minimal, custom set of messages based on the objects and interactions shown in \cref{fig:lead_som} to exchange simulation information and implements a straightforward time management scheme.

\subsection{Agent} \label{sec:agent}
The agent section illustrated in \cref{fig:lead} comprises
\begin{enumerate*}[label=(\arabic*), itemjoin*={, and\ }]
    \item the policy which controls a CGF within the simulation system
    \item the learning algorithm housing the mechanisms needed for the agent to learn to solve specific tasks
    \item the optimizer for the hyperparameters of the learning algorithm.
\end{enumerate*}
During learning, the selected ML algorithm uses the agent's states, actions and possibly rewards to improve its policy and perform the given task better. The agent can also operate without the learning algorithm, in which case the policy function yields new actions as before, but the resulting state and reward are not used to refine the policy function.

The agent interacts with LEAD through the Gymnasium interface, which defines a set of abstract methods implemented in LEAD. The \texttt{step()} and \texttt{reset()} methods are of particular relevance. The \texttt{step()} method, which takes an action as argument, causes LEAD to advance one step and returns the new state and reward of the agent and a termination flag, as illustrated in \cref{fig:lead}. The \texttt{reset()} method, which takes no argument, terminates the current episode, starts a new episode, and returns the initial state of the agent in the new episode. 

The ML algorithm may either be an RL algorithm or an IL algorithm. Reinforcement learning algorithms use states, actions and rewards collected using the agent's latest policy (sometimes combined with data generated by earlier versions of the policy) to optimize the policy to maximize future rewards. In contrast, IL algorithms use prerecorded sets of states and actions to optimize the agent policy.

Regardless of the learning type, hyperparameters specific to the learning algorithm must be set manually or by using a hyperparameter optimizer such as Optuna \shortcite{akiba19}. While policies are being optimized, the user may monitor their progression through logs and plots. Monitoring can help determine whether the learning progresses as expected or whether to apply changes to reward functions, state extractors, simulation termination criteria, or hyperparameters. Currently, these plots are created using logs generated by helper functions of the applied ML algorithms, such as those found in Stable Baselines3.

\section{EXPERIMENT} \label{sec:experiment}
This section describes an experiment aimed at learning pilot-specific behavior using RL, demonstrating the concept and usage of LEAD. The experiment concerned a two-ship formation flight involving a lead aircraft, a wingman aircraft, and a formation point specifying the desired position of the wingman. While the behavior of the lead aircraft was simple and predetermined, the wingman aircraft was controlled by the agent that learned by reinforcement. During the learning process, the agent was given formation points specifying desired formations relative to the lead aircraft. A reward function gave the agent rewards in line with its achievement.

\Cref{sec:experiment_config} describes how LEAD was configured for this experiment regarding the agent's states, actions, task, goal and reward function. \Cref{sec:experiment_ppo} reviews the proximal policy optimization (PPO) algorithm used to optimize the agent's policy. The results of the experiment are presented in \cref{sec:experiment_results}.

\subsection{Configuration of LEAD} \label{sec:experiment_config}
This section gives the configuration of LEAD corresponding to the blue section in \cref{fig:lead}. The aircraft were simulated in SACS, connected to the simulation interpreter and agent gateway using the ML connection, allowing the agent to observe the environment and control the wingman.
For simplicity, the altitude controls were held constant at \SI{5000}{m}, such that the geometry in reality was two-dimensional. Furthermore, the lead aircraft was set to fly in a straight line.
\begin{figure}[bt]
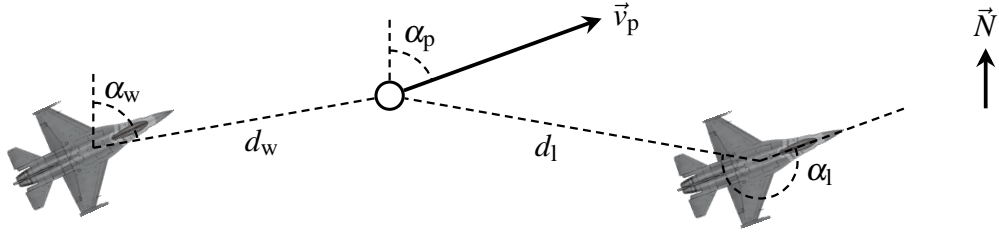

    \centering
    \include{tikz/formation}
    \vspace*{-7mm}
    \caption{Formation flight with a lead aircraft (right) and wingman aircraft (left). North is up. The formation point (black circle) represents where the wingman is instructed to fly, given by a certain distance $d\_l$ and aspect angle $\alpha\_l$ from the lead aircraft. However, the wingman deviates by a distance $d\_w$ with a bearing angle $\alpha\_w$. The formation point moves according to the lead aircraft with speed $v\_p$ at an angle $\alpha\_p$.
    \linebreak Aircraft image source: Goldhawk Interactive.}
    \label{fig:formation}
\end{figure}

A single goal described the task of the wingman agent: to achieve and maintain the desired formation. The formation point was defined by a clockwise angle $\alpha\_l$ from the lead aircraft direction and a distance $d\_l$, as shown in \cref{fig:formation}. The agent's actions set the desired course $\alpha\_d$ and speed $v\_d$ for the wingman aircraft. All action variables and state variables are summarized in \cref{tab:actions_observations}.
\begin{table}
\centering
\caption{The variables that make up the actions and states of the agent, with symbol, domain and unit.}
\begin{tabular}{lc@{}D{,}{,\,}{4.4}S}
\toprule
Action variables                         & Symbol     & \multicolumn{1}{c}{Domain} & \text{Unit} \\\midrule
Desired course                           & $\alpha\_d$ & [0, 2\pi]                 & \si{rad} \\
Desired speed                            & $v\_d$      & [100, 500]                & m/s         \\[1.5ex]
State variables &&&\\\midrule
Course of formation point                & $\alpha\_p$ & [0, 360]                  & \si{deg} \\
Speed of formation point                 & $v\_p$      & [200, 400]                & m/s        \\
Bearing from wingman to formation point  & $\alpha\_w$ & [0, 360]                  & \si{deg} \\
Distance from wingman to formation point & $d\_w$      & [0, \infty)               & m          \\
\bottomrule
\end{tabular}
\label{tab:actions_observations}
\end{table}

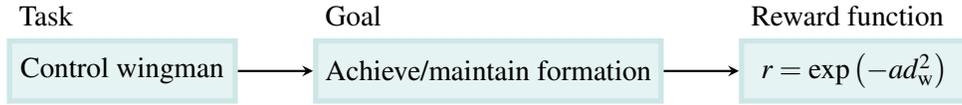
\begin{figure}
	\centering
    \begin{tikzpicture}[every node/.style={minimum height=.8cm, minimum width=3cm, bg=Green}, every path/.style={-stealth, thick}]
	\node (tsk) {Control wingman};
	\node [above=0 of tsk.north west, lab] {Task};
	\node (go1) [right=1 of tsk] {Achieve/maintain formation};
	\node [above=0 of go1.north west, lab] {Goal};
	\node (rw1) [right=1 of go1.north east, anchor=north west] {$r=\exp\left(-ad\_w^2\right)$};
	\node [above=0 of rw1.north west, lab] {Reward function};
	\draw (tsk.east) -- (go1.west);
	\draw (go1.east) -- (go1 -| rw1.west);
\end{tikzpicture}
	\caption{The task of the agent is to control the wingman aircraft, which amounts to achieving and maintaining formation, measured by a Gaussian reward function with $a=$ \SI{5e-7}{m^{-2}}.}
	\label{fig:task}
\end{figure}

\Cref{fig:task} illustrates how the task is broken down into a goal and a reward function. At each time step during learning, the agent is rewarded with a value $r\in[0, 1]$ in proportion to its proximity to the formation point. The idea behind using a continuously decreasing function without a hard cutoff is to judge the agent's behavior even when the distance to the formation point is large. Conceivably, this measure improves convergence early in the learning phase, regardless of the initial position of the wingman aircraft. The parameter $a$ determines the decay of the reward function. The larger the value $a$, the closer the wingman must get to receive substantial rewards.

All episodes were terminated after \SI{360}{s} simulated time. However, in the early phase of learning the lead and wingman aircraft would sometimes move away from each other indefinitely. Data collected in these situations contributed little to learning, so episodes were also terminated if every reward was less than \num{1e-9} over a period of \SI{180}{s}.

Exposing the agent to many different situations in the environment should help it learn how to reach the formation position in a general fashion. Both entities were therefore assigned random initial positions for every new episode. The formation point and the course and speed of the lead aircraft were constant throughout an episode, while the agent commanded the wingman aircraft with new actions every \SI{1}{s}.

\subsection{Learning Formation Flight with Proximal Policy Optimization} \label{sec:experiment_ppo}
Proximal policy optimization (PPO) was the RL algorithm of choice due to its promising results when applied to a variety of other games and tasks \shortcite{schulman17,berner19}, including learning air combat tasks \shortcite{zhang20,piao20,li22,zhang22}. A brief description of PPO follows here, while a full description is found in \shortciteN{schulman17}. The algorithm is an actor-critic method in which the policy and the value function are represented as neural networks with parameters $\theta$. The policy and value function networks may be separate or partly connected. Partly connected networks share the input layer and one or more hidden layers.

The agent interacts with LEAD for $n$ steps using its current policy. Then, PPO updates the policy and value function networks using data samples collected from these $n$ steps. This process is repeated until the performance is satisfactory. One update consists of a series of gradient ascent steps, each adjusting the network parameters in the direction that is expected to maximize future rewards. The $n$ samples are divided into minibatches of size $m$. For each minibatch, the averaged optimization objective 
\begin{equation}
    L(\theta) = \tfrac1m\sum_{i=1}^m\big(L\*{clip}(\epsilon, \gamma, \theta)+cS(\theta) - L\*{VF}(\theta) \big),
\label{eq:ppo_objective}
\end{equation}
is used to perform one gradient ascent step, with an update step size given by the learning rate. All hyperparameters are summarized in \cref{tab:ppo_hyperparameters}. The objective in \eqref{eq:ppo_objective} consists of three terms which we describe here only through their effect on the learning process. The first term $L\*{clip}(\epsilon, \gamma, \theta)$ estimates the quality of the actions of the policy, where a number above one is good. Further, $L\*{clip}$ is clipped to the interval $[1-\epsilon, 1+\epsilon]$ to make the quality estimates more conservative. The second term includes the entropy $S$ of the policy, which is higher for policies without a strong preference for any particular actions in a given state. Thus, the entropy term awards exploration of new behavior strategies. The role of the entropy term scales with a constant $c$ and will diminish with time as $L\*{clip}$ increases. The third term in \eqref{eq:ppo_objective} is used to update the parameters of the value function network, which are updated in parallel with the policy parameters. The value function is an estimator of future rewards based on the state, and it is used by $L\*{clip}$.
Each gradient ascent step over a set of $n/m$ minibatches is called an \emph{epoch}. It is common to run multiple epochs per policy update and to shuffle the samples among the minibatches between each epoch.
\begin{table}
\centering
\caption{The main hyperparameters of PPO in the Stable Baselines3 implementation. The optimized hyperparameter values were found by Optuna by running 200 trials for \num{30000} learning steps.}
\begin{tabular}{TCD{,}{,\,}{4.6}cl@{}}
\toprule
\textnormal{Hyperparameter} & \text{Symbol} & \multicolumn{1}{c}{\makecell{Search\\boundaries}} & \makecell{Optimized\\value}  & Description  \\ 
\midrule
batch\textunderscore size    & m        & [16, 1024]                          & 64            & Size of minibatch \\
learning\textunderscore rate &          & \big[\num{1e-4}, \num{1e-1}\big]    & \num{1.3e-3}  & Step size of gradient update \\
n\textunderscore steps       & n        & [1024, 4096]                        & 2048          & Number of steps to run per update \\
n\textunderscore epochs      &          & [3, 50]                             & 47            & Number of epochs \\
gamma          & \gamma   & [0.9, 0.9999]                       & 0.9467        & Discount for long-term rewards \\
clip\textunderscore range    & \epsilon & [0.1, 0.4]                          & 0.1359        & Clipping range for action values \\
ent\textunderscore coef      & c        & \big[\num{1e-8}, \num{1e-1}\big]    & \num{5e-4}    & Entropy coefficient in objective \\
\bottomrule
\end{tabular}
\label{tab:ppo_hyperparameters}
\end{table}

In the experiment, the policy was as a three-layer neural network with four input nodes, a hidden layer with 64 nodes, and two output nodes, all using the hyperbolic tangent activation function. The value function was a separate neural network with a similar topology, except it had a single output node. We used the PPO implementation found in Stable Baselines3, and the primary hyperparameters of the algorithm are summarized in \cref{tab:ppo_hyperparameters}. As with most RL methods, the hyperparameter values greatly influence the performance of the algorithm \shortcite{zhang21}. However, they may be difficult to get right, so rather than hand-tuning the hyperparameters of PPO, these values were found using the optimization framework Optuna with a covariance matrix adaptation evolution strategy sampler \shortcite{hamano22}.

\subsection{Results} \label{sec:experiment_results}
\begin{figure}
    \centering
    \includegraphics[width=.8\linewidth]{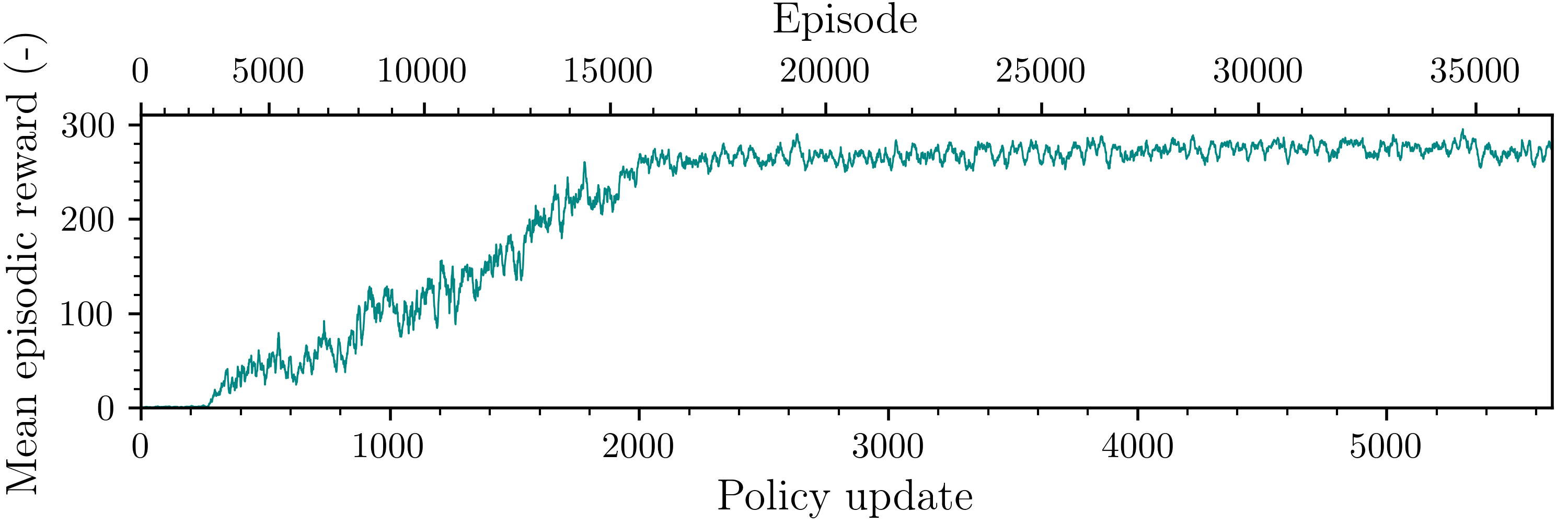}
    \vspace*{-3mm}
    \caption{Mean reward of the previous 100 episodes evaluated at each of the 5664 policy updates. The top axis is the approximate number of episodes.}
    \label{fig:reward}
\end{figure}

The agent's learning progress is measured by the local average of collected rewards. The mean episodic reward of the 100 last episodes was computed at each policy update and graphed in \cref{fig:reward}. The approximate number of episodes is given on the top axis. Until policy update $\sim$250, the agent explored actions that yielded no reward, but after this, the reward grew consistently until policy update $\sim$2200. Although the learning lasted for yet another 3000 episodes, the performance of the agent did not improve. While the absolute maximum reward per episode is 360, a value close to 300 is a realistic maximum, considering the time spent to reach the formation. The plateau at policy update 2200 corresponds to \SI{1290}{h} simulated flight time, which was \SI{4}{h}~\SI{18}{\minute} wall clock time on a computer with \SI{12}{GB} of memory and an Intel Xeon X5660 central processing unit with six cores running at \SI{2.8}{GHz}.

After learning, the agent policy was saved for further use and examination. First, the policy was validated qualitatively by visualizing the behavior. For this, we made an interface for moving the formation point in real time. Using VR-Forces, we visualized the agent's response to various formation points and concluded that the wingman agent could quickly adapt and maintain desired formations.
In order to obtain performance details, the agent was run in LEAD for 9000 episodes of \SI{600}{s} with the final policy. The starting position of the lead aircraft, wingman aircraft and formation point were randomly sampled. In 7597 out of 9000 episodes, the wingman was able to achieve $d\_w<\SI{250}{m}$ within the limit of \SI{600}{s}, and naturally, the initial distance to the formation point was a strong indicator of the time required to reach the formation. The agent made new actions every \SI{1}{s} when learning, but this interval was reduced to \SI{0.1}{s} when testing. The simulation time step was \SI{0.1}{s} in both cases.

It is worth mentioning that the abilities of the agent seemed to exceed the specified task. The agent could maintain formation most of the time even in tests where the lead aircraft made several turns, despite not seeing this while learning.

While the content in this section has concerned a single run of PPO, other runs were conducted for comparison. In fact, PPO is stochastic such that replicate runs differ to some extent. The repeatability was not quantified, but there were some qualitative takeaways after multiple runs of PPO. About half of the runs produced an agent similar to the one presented, and the reward graphs increase from policy update $\sim$250 onward. The remaining runs produced agents which seemingly had not learned anything, which may be caused by differences in the stochastic initialization of the policy network, initialization of episodes and action selection. Additionally, we attempted other configurations of LEAD, in terms of reward functions, simulation termination criteria, and time steps. For each configuration, Optuna was applied to optimize the hyperparameters of PPO. However, some of the configurations only worked after additional hand-tuning of the hyperparameters.

\section{Discussion} \label{sec:discussion}
Users of LEAD may apply many different ready-made ML algorithms due to the Gymnasium interface. This is a useful feature because it is sometimes unclear beforehand which algorithm is best suited for learning to perform a given task. Still, the user may be overwhelmed by having to configure many parameters for both LEAD and the ML algorithm. Therefore, tools for simplifying system configuration should be developed. Currently, the user may monitor the agent learning progress through logs and plotting tools based on Stable Baselines3 functionality, for instance to create graphs akin to \cref{fig:reward}. Future development may include LEAD-specific monitoring and logging tools. 

The simple air combat simulation, SACS, is fast and reliable. However, further benchmarks on different computer specifications are pending, along with scrutiny of where speedups are possible. Another planned task is a comparison of the FDM of SACS against a high-fidelity FDM.

The RL experiment described in \cref{sec:experiment} is a simple demonstration of behavior modeling capabilities LEAD offers. The agent learned using the Stable Baselines3 implementation of PPO. Unfortunately, there was no guarantee for learning success as the agent occasionally would fail to learn formation flight upon new learning runs, even when rerunning successful learning schemes. This inconsistency indicates that the success of learning crucially depends on more than the PPO hyperparameters, such as the initialization of the neural networks. Though hyperparameter values that make learning succeed every time may exist, further work with incorporating the Optuna hyperparameter optimization framework into LEAD is desirable, such as selecting appropriate search boundaries (\cref{tab:ppo_hyperparameters}) and value samplers.
Additionally, it seems possible to accelerate the learning by reducing the time when the wingman aircraft receives almost no reward by either flattening the reward function, initializing the aircraft closer to each other or ending failed episodes sooner. However, the agent must also learn to recover when it is far from the goal.

It took just over four hours to train the wingman agent, which is a considerable but acceptable amount of time. The bottleneck does not appear to be SACS, but rather the data-intensive nature of RL algorithms, including PPO. However, the entire learning process is currently sequential. The next natural step is therefore to parallelize simulations in LEAD. With Gymnasium's built-in support for vectorizing environments, LEAD can be parallelized using software containers and orchestration systems. Larger batch sizes can then be used to update policies using graphics processing units and presumably speed up the learning process.

Future work may include attempts at learning more complex tasks and scenarios such as multi-agent configurations, combat engagements and the use of weapons. Imitation learning is applicable for these learning tasks to ensure that learned policies not only complete tasks but also comply with demonstrations and realistic behavior \shortcite{gorton22,sandstrom22}.

Future work may also investigate transfer learning with LEAD. Using the HLA connection, agents may thus learn to solve a task using SACS, then attempt to solve the task in another simulation system. Another possibility is for agents to learn across several simulation systems of different fidelity.

\section{CONCLUSION} \label{sec:conclusion}
The future of simulation-based training for fighter pilots is expected to involve pilots engaging in combat with intelligent and realistically behaving CGFs. To explore the potential of artificial intelligence methods applied to this context, LEAD was developed, a system for creating behavior models for agents operating in the air domain. The system is fully configurable and designed to be interoperable with third-party simulation software using distributed simulation. It incorporates the Gymnasium interface developed by OpenAI, making LEAD easy to use with popular programming libraries that implement modern RL and IL algorithms.

The usage and performance of LEAD was demonstrated in an experiment where an agent successfully learned formation flight using RL. Configuring LEAD for the experiment was easy due to the modular system design, allowing users to add new functionality when needed. While the agent was learning, it was easy to monitor its progress, particularly through logs and reward plots.

About four hours into the experiment, corresponding to \SI{1290}{h} simulated flight time, the agent was proficient and attained any assigned formation with few mistakes. While flying in formation may be considered a simple task, the capability of learning is clear and appears directly extensible to more complex tasks. A qualitative assessment revealed that the agent even attained formation when the lead aircraft changed course regularly despite not having experienced those situations during learning. The ML experiment was made as simple as possible by keeping the number of active entity subsystems to a minimum. Hence, it remains to conduct experiments where the agent learns to use weapons and communicate with other entities.

The experiment also served as an initial test of the Simple Air Combat Simulation. There is still a need to validate the dynamics model properly. Future improvements to SACS aim to increase the simulation speed and facilitate parallel execution to enhance the learning of agents.
Thus far, distributed simulation is used only to visualize aircraft simulated by SACS, primarily to evaluate the agent performance. It remains to run experiments where a third-party simulation system replaces SACS to investigate to what extent behaviors learned using LEAD generalize across environments and tasks.

While LEAD has already proved convenient and effective in producing agent behavior for an air-based CGF, the experiences made and the flaws unveiled by the demonstration will help improve LEAD for new research. Future experiments will include other learning algorithms and explore the capabilities of LEAD for learning realistic air combat behavior beneficial for simulation-based training of fighter pilots.

\footnotesize 
\bibliographystyle{wsc}
\bibliography{wsc}

\section*{AUTHOR BIOGRAPHIES}

\noindent {\bf ANDREAS STRAND} \orcidlink{0000-0002-7679-7816} is a Scientist at FFI (Norwegian Defence Research Establishment). He holds a PhD in quantification of uncertainty in simulations. Currently, his primary research interests are intelligent agents for air combat simulations and decision support for military operations. His email address is \email{andreas.strand@ffi.no}.\\

\noindent {\bf PATRICK R. GORTON} \orcidlink{0000-0001-8588-1764} is a Scientist in the Simulation and training research program at FFI. His primary research interests include artificial intelligence, machine learning, and modeling and simulation, with focus on behavior modeling of intelligent agents for military training and decision support. His email address is \email{patrick-ribu.gorton@ffi.no}.\\

\noindent {\bf MARTIN ASPRUSTEN} is a Senior Scientist at FFI in the field of distributed simulation, modeling and system development for military applications. His email address is \email{martin.asprusten@ffi.no}.\\

\noindent {\bf KARSTEN BRATHEN} is a Chief Scientist at FFI. His research interest is in the area of AI based simulation support to operation planning and tactical training. His email address is \email{karsten.brathen@ffi.no}.
\end{document}